\documentclass{article}
\usepackage{amsmath,graphicx,mlspconf}

\title{Enhancing Supervised Visualization through Autoencoder and Random Forest Proximities for Out-of-Sample Extension}
\name{%
   Shuang Ni$^{\star \dagger}$%
   \qquad Adrien Aumon$^{\star \dagger}$%
   \qquad Guy Wolf$^{\hspace{.2em} \star \dagger}$%
   \qquad Kevin R. Moon$^{\ddagger}$%
   \qquad Jake S. Rhodes$^{\mathparagraph}$.
}%

\address{%
   $^{\star}$ Department of Mathematics and Statistics, Université de Montréal \\%
   $^{\dagger}$ Mila - Quebec AI Institute \\%
   $^{\ddagger}$ Utah State University \\%
   $^{\mathparagraph}$ Brigham Young University
}

\begin{document}

\maketitle

\begin{abstract}
The value of supervised dimensionality reduction lies in its ability to uncover meaningful connections between data features and labels. Common dimensionality reduction methods embed a set of fixed, latent points, but are not capable of generalizing to an unseen test set. 
In this paper, we provide an out-of-sample extension method for the random forest-based supervised dimensionality reduction method, RF-PHATE, combining information learned from the random forest model with the function-learning capabilities of autoencoders. Through quantitative assessment of various autoencoder architectures, we identify that networks that reconstruct random forest proximities are more robust for the embedding extension problem. Furthermore, by leveraging proximity-based prototypes, we achieve a 40\% reduction in training time without compromising extension quality. Our method does not require label information for out-of-sample points, thus serving as a semi-supervised method, and can achieve consistent quality using only 10\% of the training data.
\end{abstract}
\section{Introduction}
\label{sec:intro}
In the current era of high-dimensional data, traditional univariate and bivariate visualization tools often fail to accurately represent variable relationships. Dimensionality reduction methods address this challenge by embedding data into lower-dimensional representation, with the aim of revealing intrinsic patterns crucial for understanding complex datasets. This can be accomplished using modern manifold learning methods.

Manifold learning methods, such as Isomap~\cite{Tenenbaum2000isomap}, Locally Linear Embedding~\cite{rowies2000lle}, and Diffusion Maps~\cite{nadler2006diffusion}, have introduced the concept of data manifolds, which represent the intrinsic low-dimensional geometry within high-dimensional data. However, such methods often fall short for visualization tasks due to their tendency to embed data into different dimensions~\cite{Haghverdi2016dpt}. To address this limitation, methods like $t$-SNE~\cite{vanDerMaaten2008tsne}, UMAP~\cite{mcinnes2018umap}, and PHATE~\cite{Moon2019phate} have emerged and offer more suitable solutions for visualizing complex datasets.

In several domains, experts provide labels that can serve as auxiliary information. Advances in supervised machine learning have shown that computers can provide valuable insights into the relationships between features and label information. Traditional manifold learning techniques typically do not account for this label information; however, such insights can be tailored to fit the manifold learning framework. Examples of this include supervised variants of $t$-SNE\cite{cheng2015stsne,hajderanj2019stsne}, Isomap~\cite{Ribeiro2008esiso}, and UMAP~\cite{sainburg2021ssumap}. 

More recently, the authors of~\cite{rhodes2024gaining} demonstrated the weaknesses of these supervised methods (e.g., hyper-separation, undesirable embeddings for noisy
data) and proposed an alternative approach, based on random forest (RF) proximities~\cite{Breiman2001randomforests} and PHATE~\cite{Moon2019phate}. This method, RF-PHATE, is capable of embedding meaningful relationships in a variety of noisy data spaces, such as Raman spectroscopy and COVID-19 patient plasma data~\cite{rhodes2024gaining}. 
Visualization using RF-PHATE also helped to identify a less understood subclass of multiple sclerosis~\cite{rhodes2024gaining}.

Most manifold learning algorithms (both supervised and unsupervised) generate fixed coordinates within a latent space, but lack a mechanism to accommodate new observations. In other words, to embed previously unseen data, the algorithm must be rerun with the new data as part of the training set. Moreover, methods that require a pairwise distance matrix can be computationally prohibitive when dealing with large datasets. One well-known solution to the lack of out-of-sample extension capability is the Nystr{\"o}m extension~\cite{bengio2003out} and its variants, such as geometric harmonics~\cite{coifman2006geometric}. 
These methods approximate an empirical function over new points using linear combinations of the eigenvectors of a kernel matrix computed on the training set. Although this method extends to eigenvectors, its direct applicability to diffusion-based methods is limited. As an alternative extension method, recent approaches such as Geometry-Regularized Autoencoders (GRAE)~\cite{duque2022geometry}, offer a promising solution for extending embeddings to out-of-sample data points. 


In this paper, we propose a novel extension of RF-PHATE aimed at effectively handling out-of-sample instances adapted for semi-supervised dimensionality reduction. We introduce a set of regularized autoencoder (AE) architectures that address out-of-sample extension, taking inspiration from the principles of GRAE~\cite{duque2022geometry} which uses a manifold embedding to regularize the AE. In addition to this regularization, we also incorporate the supervised information based on RF proximities, enabling us to learn an embedding function extendable to new data in a supervised context. We empirically show that AE methods that reconstruct pairwise proximities are better suited to extend embeddings than AEs that reconstruct the original data. Furthermore, we introduce a proximity-based landmark selection approach to augment our method and reduce training time by 40\%. The extended embedding is comparable to the full embedding even when constructed using only 10\% of the training data. Notably, our approach extends out-of-sample examples without relying on label information for new points, which is helpful in situations where label information is expensive or scarce. This method improves the adaptability and scalability of the manifold learning process, allowing for seamless integration of new data points while maintaining the desirable traits of established embedding methods.



\section{Background}
\label{sec:Background}
We complement RF-PHATE's visualization strengths with the extensibility of AEs to enhance this supervised embedding method. Our method maintains the visual quality of RF-PHATE while providing a natural approach to out-of-sample extension via an AE. We employ auxiliary label information in the architecture of the AE, allowing us to take advantage of the insights of experts and retain the supervised model learning throughout the visualization function-learning process. Our experiments show that the RF-PHATE AE extensions can effectively construct a latent representation that is faithful to the original RF-PHATE algorithm.

\subsection{Geometry Regularized Autoencoders}
\label{subsec:GRAE}
AE networks serve a dual purpose: first, they learn an encoder function $f_e(X)$ (where $X$ can be tabular data, images, etc.), which compresses the input data into a latent representation via a bottleneck layer~\cite{theis2017lossy}. Second, they learn a decoder function, $f_d(Z)$, that aims to map this lower-dimensional embedding back to the original input space. In other words, AEs seek to find an optimal set of encoding and decoding functions such that $f_d(f_e(X)) \approx X$, by minimizing a loss function $L(f_e, f_d)$. The loss function measures the ability of the AE to reconstruct the initial input. Here, we denote the reconstruction loss as $L(f_e, f_d) = L_{recon}(X, f_d(f_e(X)))$, where $L_{recon}(., .)$ is the mean squared error (MSE).

Through the loss minimization process, the AE learns to represent data in a more compact form, while simultaneous training of the decoder function ensures that the latent representation is meaningfully related to the original data. However, the standard AE's encoded representation often struggles to accurately capture the underlying data geometry and the representation does not align with human-interpretable visualizations~\cite{duque2022geometry}. To ensure the learned latent-space function maps to a space capable of aiding humans in the task of exploratory data analysis, we require the encoder to produce semantically significant embeddings by ensuring the AE embeddings do not differ significantly from the RF-PHATE embeddings. 

Given a learned RF-PHATE embedding, $G$, we force the AE to learn a representation similar to $G$ by regularizing the loss function at the bottleneck layer as follows: $L(f_e, f_d) = L_{recon}(X, f_d(f_e(X))) + \lambda L_{geom}(f_e(X), G)$. $L_{geom}$ discourages the encoder function, $f_e$, from learning a latent representation that significantly deviates from the embedding $G$. The parameter $\lambda$ controls the degree to which the embedding is used in encoding $X$. This approach not only estimates a manifold extendable to new data points but also able to reconstruct data from embeddings, providing benefits in terms of data reconstruction and data generation.

To leverage the knowledge gained from an RF model, we modify the AE architecture (Section~\ref{sec:oose}) to further incorporate the RF's learning. The forest-generated proximity measures~\cite{rhodes2023rfgap}, which indicate a similarity between data points relative to the supervised task, serve as a foundation for extending the embedding while integrating the insights acquired through the RF's learning process. Both new and existing data can be effectively incorporated into the learned embedding space while maintaining the integrity of the underlying manifold structure and without retraining either the forest or rerunning the full RF-PHATE embedding algorithm. 

\subsection{Random Forest Proximities and RF-PHATE}
\label{subsec:RFProx}

Random forests~\cite{Breiman2001randomforests} consist of an ensemble of randomized decision trees, each of which partitions a bootstrap sample of the training data. Observations within the bootstrap sample are termed ``in-bag", while those in the training data but not included in the sample are referred to as ``out-of-bag" (OOB). These partitions create a decision space for classification or regression. This decision space naturally forms boundaries that can be used to extract information regarding observational similarity. 

Naively, the similarity measure or proximity between two observations, $\mathbf{x}_i$ and $\mathbf{x}_j$, denoted as $p(\mathbf{x}_i, \mathbf{x}_j)$, can be defined as the number of terminal nodes they share across all trees normalized by the size of forest~\cite{Breiman2001randomforests}. However, this approach overemphasizes class segregation. To address this issue, an alternative formulation was developed to calculate pair-wise similarities only between observations that are OOB~\cite{hastie2017elements}. Despite reducing overinflated class separation, OOB-only proximities tend to to be sensitive to noise~\cite{rhodes2023rfgap}. 

In~\cite{rhodes2023rfgap}, a new proximity formulation was introduced to address these limitations. This formulation is called Random Forest-Geometry- and Accuracy-Preserving proximities (RF-GAP). RF-GAP proximities are specifically designed to mimic the RF decision function when using a proximity-weighted neighbor classifier or regressor. In doing so, they capture the relevant information for the prediction problem. RF-GAP proximities were shown to improve various applications, including data imputation, outlier detection, visualization via multidimensional scaling (MDS), and enhanced random forest local interpretability~\cite{rosaler2023enhanced}.

Although not discussed in~\cite{rhodes2023rfgap}, we describe how RF-GAP proximities can be extended to a new test set or out-of-sample points, which will be needed for our autoencoder regularization. Each out-of-sample observation can be considered OOB. Thus, for an out-of-sample point, $\mathbf{x}_o$, we calculate:
\[p_{GAP}(\mathbf{x}_0, \mathbf{x}_j)=\dfrac{1}{|T|} \sum_{t \in T}  \frac{c_j(t) \cdot I\left(j \in J_{0}(t)\right)}{\left|M_{0}(t)\right|},\]
where $T$ is the set of all trees in the forest of size $|T|$, $c_j(t)$ is the number of repeated samplings for observation $\mathbf{x}_j$ in tree $t$, $I(\cdot)$ is the indicator function, $J_0(t)$ is the set of in-bag points residing in the terminal node of observation $\mathbf{x}_0$ in tree $t$, and $M_0(t)$ is the multiset of in-bag indices, including repetitions.




The RF-GAP definition requires that self-similarity be zero, that is, $p_{GAP}(\mathbf{x}_i, \mathbf{x}_i) = 0$. However, this is not suitable as a similarity measure in some applications. Due to the scale of the proximities (the rows sum to one, so the proximity values are all near zero for larger datasets), it is not practical to simply re-assign self-similarities to 1. Otherwise, self-similarity would carry equal weight to the combined significance of all other similarities. Instead, we assign values by, in essence, passing down an identical OOB point to all trees where the given observation is in-bag. That is:

\[p_{GAP}(\mathbf{x}_i, \mathbf{x}_i)=\dfrac{1}{\left|\bar{S}_{i}\right|} \sum_{t \in \bar{S}_{i}}  \frac{c_i(t)}{\left|M_{i}(t)\right|},\]
where $\left|\bar{S}_{i}\right|$ is the set of trees for which $\mathbf{x}_i$ is in-bag. This formulation guarantees that $p_{GAP}(\mathbf{x}_i, \mathbf{x}_i) \ge p_{GAP}(\mathbf{x}_i, \mathbf{x}_j)$ for all $j \ne i$. Additionally, $p_{GAP}(\mathbf{x}_i, \mathbf{x}_i)$ under this formulation is on a scale more similar to other proximity values.


RF-PHATE employs RF-GAP proximities to represent supervised similarity, from which the global structure is eventually learned through a diffusion process. The proximities encompass a hierarchical learning paradigm that contains information about important supervising data features. The local and global structure representation is eventually encoded into lower dimensions through multidimensional scaling~\cite{rhodes2024gaining}.

\section{RF-PHATE Out-of-Sample Extension}
\label{sec:oose}

We denote $\mathbf{x}_E$ as an element of the extended set $\mathcal{X}_E$ (not included in the training set). For every data point $\mathbf{x}$ in the dataset $\mathcal{X}$, we generate a set of proximities, $\left\{\tilde{p}_{GAP}(\mathbf{x}_E, \mathbf{x}) \right\}_{\mathbf{x}\in \mathcal{X}}$ using a trained RF model. From the RF proximities, we create an RF-PHATE embedding, $G$, which is used to guide the AE embedding via the geometric loss for semantically meaningful visualization.

Through experiments, we have determined that incorporating the proximities in the reconstruction process leads to embeddings that are truer to the original RF-PHATE embedding, evaluated using Mantel test's~\cite{mantel1967mantel}, a statistical measure for evaluating the rank-based similarity between two distance matrices. We call the model RF-PRN (random forest proximity reconstruction network). 

\begin{figure}[t]
\centering
\includegraphics[width=0.49\textwidth]{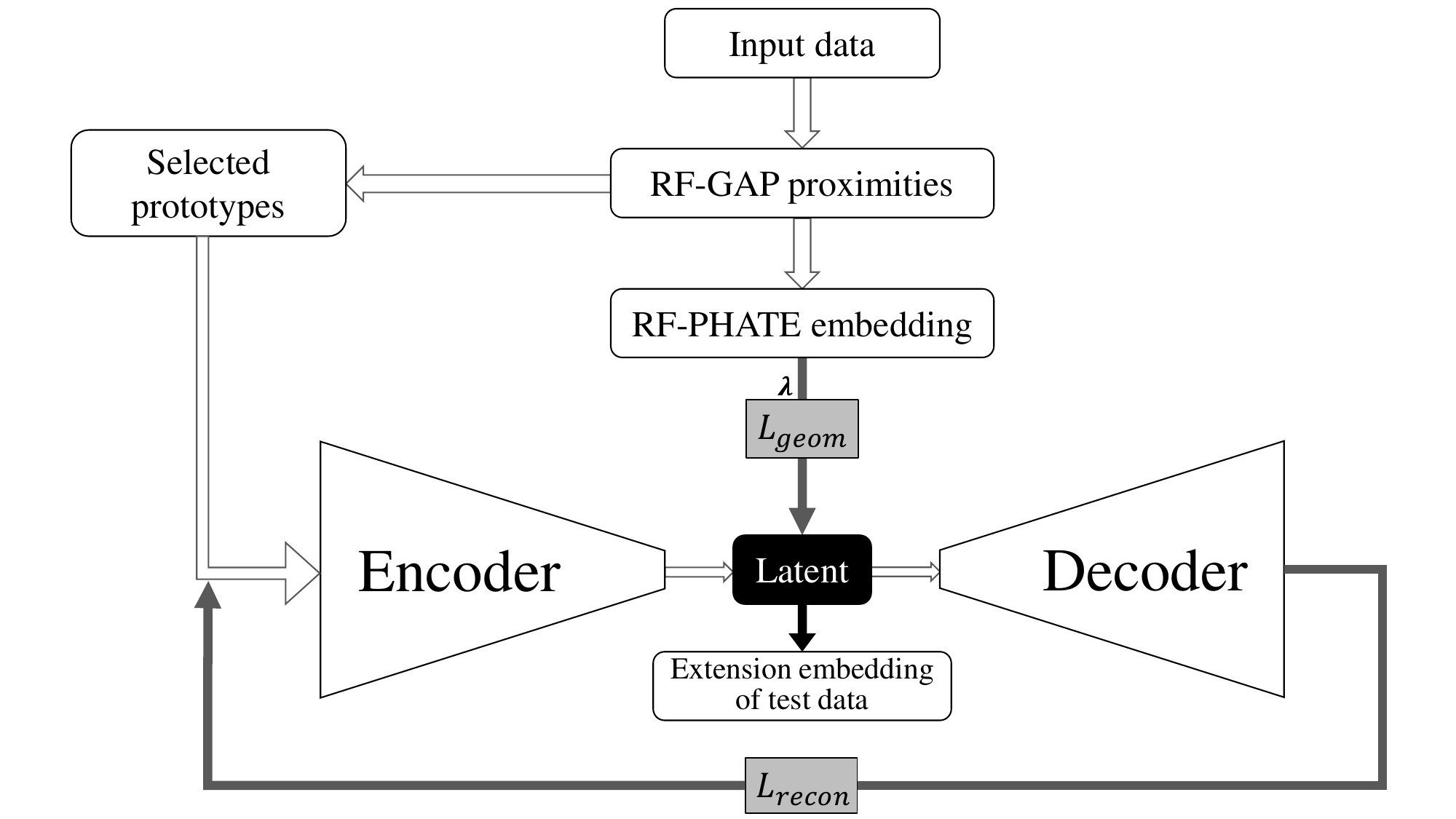}
\caption{A visual depiction of the RF-PRN-PRO architecture. Using prototypical RF-GAP proximities from the training sample, the loss function combines MSE reconstruction loss $L_{recon}$ and a geometric regularization term $L_{geom}$ involving MSE between latent representation and RF-PHATE embeddings. For out-of-sample extension, this architecture incorporates new data and proximities, using the latent representation as extended embeddings.} 
\label{fig:architecture}
\end{figure}

However, it becomes impractical to reconstruct the complete set of proximities when training datasets are large. To address this challenge, we have introduced a proximity-based landmark selection approach. This approach identifies observations that serve as representative cases of classes in the dataset, where landmarks or prototypes are chosen as the observation corresponding to the highest average within-class proximities. We call this model RF-PRN-PRO. This streamlined approach not only expedites the training process but also generates an embedding that exhibits more prototypical class representations. For a visual representative of the RF-PRN-PRO architecture, see Figure~\ref{fig:architecture}.


We have explored other adaptations to the traditional AE to optimize knowledge retention from the RF model within the autoencoder. Each is geared towards improving the quality of the extended embedding by leveraging some aspect of the RF's learning. All architectures use the geometric regularization at the bottleneck. To clarify the model inputs, we give the encode input and decoder outputs for each model, using $\mathcal{X}$ for the original data and $\mathcal{P}$ for the RF proximities.

\begin{enumerate}
    \item RF-GRAE: (\textbf{Input}: $\mathcal{X}$, \textbf{Output}: $\mathcal{X}$) This architecture matches the GRAE~\cite{duque2022geometry} model while employing the RF-PHATE embedding as a supervised regularizing embedding.
    \item RF-PROX-IN: (\textbf{Input}: $\mathcal{P}$, \textbf{Output}: $\mathcal{X}$) Here we use the RF-GAP proximities as input to the initial layer of the encoder with the aim of constructing the original data via the decoder as a type of inverse model. This feeds supervised information to the encoder learned from the RF.
    \item RF-PROX-REG: (\textbf{Input}: $\mathcal{X}$, \textbf{Output}: $\mathcal{X}$) This autoencoder follows the pattern of GRAE but has an additional level of regularization. In addition to the geometric regularization, a linear layer is added to the bottleneck aimed at predicting the proximities directly from the low-dimensional representation. This augmentation attempts to facilitate the model's capacity to not only reconstruct the input data but also directly predict the random forest proximities to the training points at the bottleneck layer.
    \item RF-PRN: (\textbf{Input}: $\mathcal{P}$, \textbf{Output}: $\mathcal{P}$) This AE takes the proximities (from the training data) as input and the decoder attempts to reconstruct the proximity values. Proximities between training and test (out-of-sample) points are thus estimated by the decoder.
    \item RF-PRN-PRO: (\textbf{Input}: $\mathcal{P}$, \textbf{Output}: $\mathcal{P}$) This architecture follows the same framework as RF-PRN, however, we take only a prototypical sample of training indices for the proximity reconstruction. A visual representation of this architecture can be found in Figure~\ref{fig:architecture}. 
\end{enumerate}

\section{Experiments}
\label{sec:experiments}
We compared the AE architectures using the Mantel test to assess how well each model extends the RF-PHATE embedding to a test set. Specifically, in our context, we generated full RF-PHATE embeddings (including both training and test data) and compute the Mantel correlation between the pairwise distances of the RF-PHATE embedding (test points only) and the pairwise distances of the AE test embedding points. A high Mantel correlation coefficient indicates a strong correspondence between the structures of the original and extended embeddings, suggesting that the extended embedding effectively preserves the underlying data relationships captured by the RF-PHATE embedding. 

Table~\ref{tab:mantel} provides the Mantel correlations between the original and extended embeddings for each architecture. These scores are aggregated across 26 publicly available datasets\footnote{From UCI~\cite{UCI2019}, or otherwise publicly available: Audiology, Balance Scale, Breast Cancer, Car, Chess, CRX, Diabetes, Flare1, Glass, Heart Disease, Heart Failure, Hepatitis, Hill Valley, Ionosphere, Iris, Lymphography, MNIST (test set), Optical Digits, Parkinson's, Seeds, Segmentation, Tic-tac-toe, Titanic, Artificial Tree Data~\cite{Moon2019phate}, Waveform, Wine.} and 10 repetitions. For each dataset, we randomly assign distinct training and test sets, using 70\% of the data to train each model and extend the embedding to the remaining 30\% using the AEs. Each dataset and combination were evaluated using $\lambda$ values of 1, 10, and 100. Mean and standard deviation calculations were performed on the results obtained for all $\lambda$ values across datasets and combinations. 

\begin{figure}[t]
    \centering
    \includegraphics[width = 0.49\textwidth]{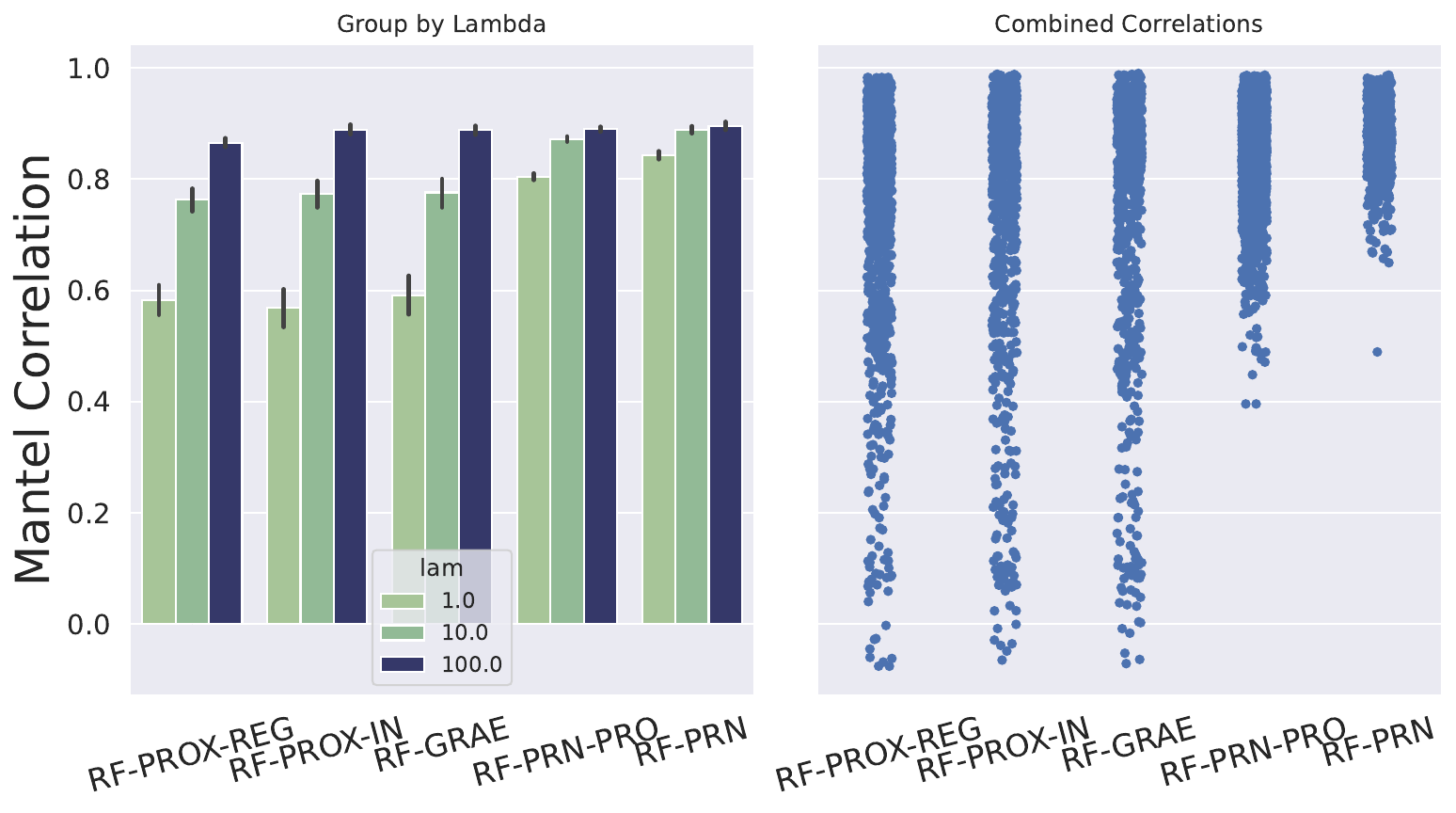}
    \caption{(Left) The mean Mantel correlations are grouped by model architecture and the regularization parameter $\lambda$. Proximity-reconstructing networks (RF-PRN and RF-PRN-PRO) tend to be more robust to the choice of $\lambda$, while the other networks give similar performance for higher $\lambda$ values. (Right) Categorical plots of each model architecture indicate that RF-PRN and RF-PRN-PRO generally produce embeddings truer to the RF-PHATE original embeddings.}
    \label{fig:mantel_comparisons}
\end{figure}

The regularized models that reconstruct proximities, rather than the original features, generally produce embedding extensions truer to the original RF-PHATE embeddings. Furthermore, proximity-reconstructing models, RF-PRN and RF-PRN-PRO, tend to be more robust to the regularization coefficient, $\lambda$, which determines the extent to which the training embedding is emphasized during the training steps, as shown in Figure~\ref{fig:mantel_comparisons}. Although models such as RF-PROX-REG, RF-PROX-IN, and RF-GRAE display similar performance for higher $\lambda$ values, RF-PRN and RF-PRN-PRO provide more consistent performance. The reconstruction of other similarity or distance measures, (perhaps unsupervised), may also improve the robustness of the embedding function, though we do not test this generalization in this paper.

\begin{figure}[t]
    \centering
    \includegraphics[width = 0.49\textwidth]{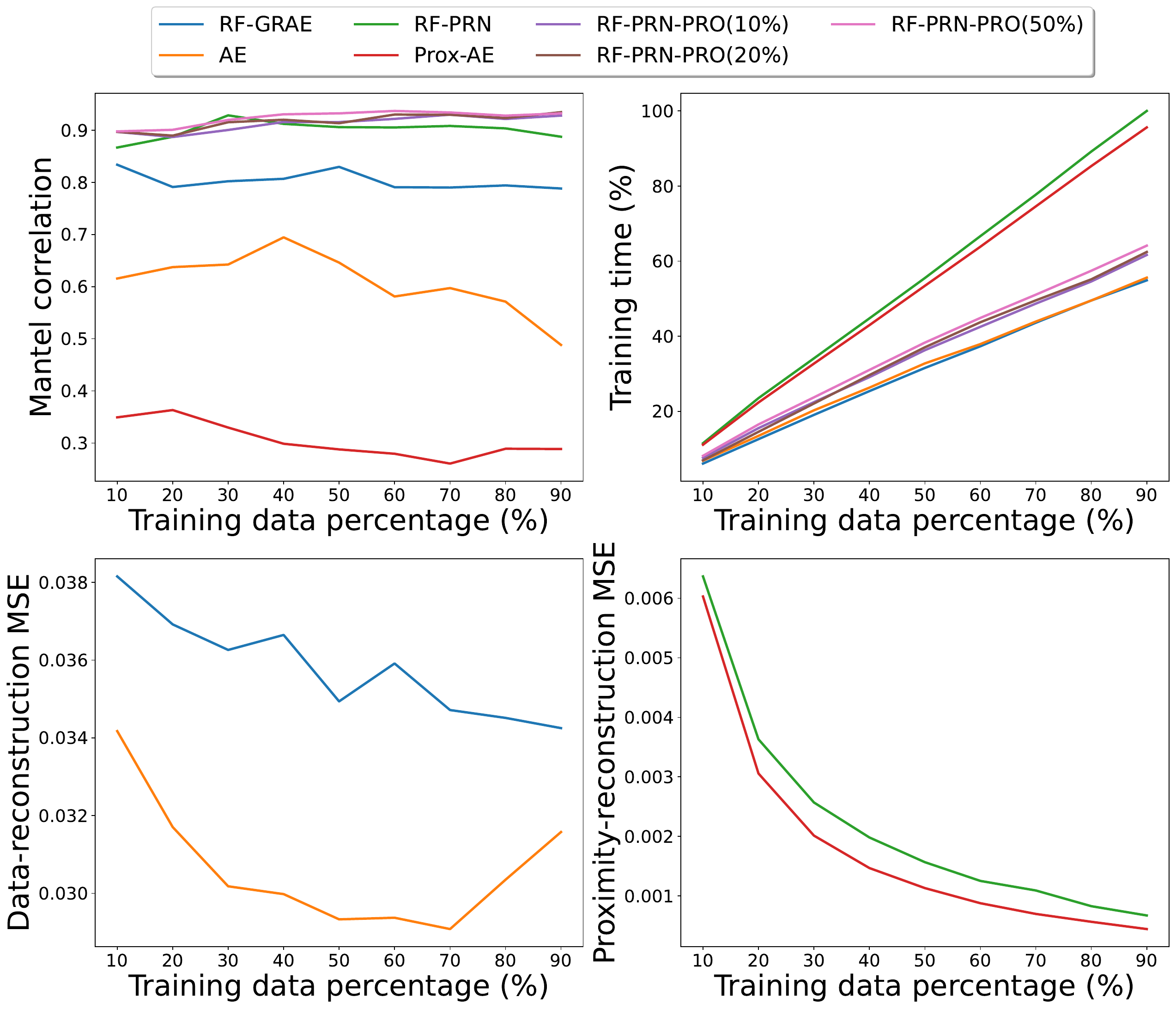}
    \caption{Evaluation of architecture performance ($\lambda = 10$) across different training data percentages using the Fashion-MNIST dataset \cite{xiao2017fashion}, with 10 repetitions. The metrics include Mantel correlation, training time, and MSE. The training time percentage is calculated as the ratio of each architecture's time to the maximum time. RF-PRN and RF-PRN-PRO consistently provide high-quality embeddings across different percentages of training data used. Although the training time for RF-PRN is higher than the other models, RF-PRN-PRO improves training time while maintaining embedding quality.}
    \label{fig:Fashion-MNIST}
\end{figure}

Due to the impracticality of reconstructing the complete set of proximities with large training datasets, RF-PRN-PRO, only uses prototypical proximity examples for reconstruction, offering a more efficient training alternative while maintaining comparable results. 
A comparison using different percentages of prototypes in RF-PRN-PRO is given in Table~\ref{tab:proto}. As expected, the efficacy of the RF-PRN-PRO model increases with the percentage of prototypes. However, even at 10\%, RF-PRN-PRO outperforms the models that do not construct proximities (see Figure~\ref{fig:Fashion-MNIST}).

\begin{table}[htb!]
    \centering
    \caption{The means and standard deviations of the Mantel correlations are averaged across the test sets of the 26 compared datasets. Results for RF-PRN-PRO were aggregated across proximity landmarks of 10, 20, and 50\%$^\dagger$. The proximity-reconstructing AEs tend to have better performance and more stability.}
    \begin{tabular}{lc}
        \hline
        Model & Mean $\pm$ Std \\
        \hline
        RF-PRN & 0.876 $\pm$ 0.06 \\
        RF-PRN-PRO$^\dagger$ & 0.856 $\pm$ 0.08 \\
        RF-GRAE & 0.752 $\pm$ 0.24 \\
        RF-PROX-REG & 0.743 $\pm$ 0.24 \\
        RF-PROX-IN & 0.742 $\pm$ 0.21 \\
        \hline
    \end{tabular}
    \label{tab:mantel}
\end{table}

\begin{table}[t]
    \centering
    \caption{Here we compare the Mantel correlation scores of RF-PRN-PRO at varying percentages of prototypical training examples used for the proximity reconstruction. A higher number of training points correlate with truer embeddings, but even at 10\%, the embeddings are better than data-reconstructing networks.}
    \begin{tabular}{lc}
    \hline
     Pct. Prototypes & Mean $\pm$ Std \\
    \hline
    10\%          & $0.766 \pm 0.071$ \\
    20\%          & $0.832 \pm 0.097$ \\
    50\%          & $0.863 \pm 0.075$ \\
    \hline
    \end{tabular}
    \label{tab:proto}
\end{table}

To assess the effects of the quantity of training data on the embedding process, we compare the quality of the extension, the reconstruction, and the fitting time at varying percentages of training points. As a baseline comparison, we introduce two additional models: a standard AE without geometric regularization, and a proximity-reconstructing AE also without regularization. The results are given in Figure~\ref{fig:Fashion-MNIST}, which plots these scores across 10 runs using the Fashion-MNIST dataset~\cite{xiao2017fashion}. A consistent held-out test set was used for validation across all training percentage levels.

Even at 10\% of the training data, RF-PRN and RF-PRN-PRO maintain Mantel correlations around 0.9, comparable to scores obtained with 90\% of the data. This suggests the proximity-reconstructing extensions are not highly sensitive to the relative size of the training set. The MSE plots of proximity-reconstructing AEs and data-reconstructing AEs are compared separately since they reconstruct different features and are thus not directly comparable. From the evaluation of MSE reconstruction loss, the AE models with regularization demonstrate comparable reconstruction performance to the AEs without regularization. In contrast, AE models without regularization are incapable of encoding the RF-PHATE embedding. The training time of RF-PRN only increases marginally compared to Prox-AE, suggesting that the regularization term imposes minimal overhead on training time. On the other hand, RF-PRN-PRO exhibits a substantial reduction in training time, as low as 60\% of the training time of RF-PRN, due to the proximity landmark selection. Leveraging only 10\% of the data for training and incorporating proximity prototypes, RF-PRN-PRO requires less than 8\% of the training time required by RF-PRN trained using 90\% of the data while maintaining comparable performance. Moreover, the independence of the extension process from label information for out-of-sample examples positions this extension as a promising solution for semi-supervised embedding of large datasets.

\section{Conclusion}
The significance of supervised dimensionality reduction lies in its ability to reveal meaningful relationships between features and label information. RF-PHATE stands out as a strong solution in the domain of supervised data visualization; however, this method inherently lacks an embedding function for out-of-sample extension. To overcome this limitation, we extended RF-PHATE by introducing five regularized AE architectures, each designed to effectively visualize out-of-sample data points. Our experimental results confirmed the utility of these AE-based extensions, illustrating their capacity to extend embeddings to new data points while preserving the intrinsic manifold structure. We found that proximity reconstruction in the AE improves the robustness of the embedding function. The RF-PRN-PRO architecture, which employs a subset of prototypes for proximity reconstruction, offers a more efficient training alternative while maintaining comparable results. By training with only a small portion of the dataset, proximity reconstruction methods substantially reduce training time compared to training a large portion of data without sacrificing performance. This is particularly advantageous for large datasets. Furthermore, the ability to extend embeddings without relying on label information for out-of-sample examples makes our approach a promising solution for semi-supervised tasks with extensive datasets.




\bibliographystyle{IEEEbib}
\bibliography{references}

\end{document}